\documentclass{article}

\usepackage{arxiv}

\usepackage[utf8]{inputenc} 
\usepackage[T1]{fontenc}    
\usepackage{hyperref}       
\usepackage{url}            
\usepackage{booktabs}       
\usepackage{nicefrac}       
\usepackage{microtype}      
\usepackage{lipsum}
\usepackage{graphicx}
\graphicspath{ {./images/} }
\usepackage{array}
\usepackage[caption=false,font=normalsize,labelfont=sf,textfont=sf]{subfig}
\usepackage{textcomp}
\usepackage{stfloats}
\usepackage{physics}
\usepackage{url}
\usepackage{xcolor}
\usepackage{verbatim}
\usepackage{amsmath,amsfonts}
\DeclareMathOperator{\argmax}{arg\,max}
\usepackage[caption=false,font=normalsize,labelfont=sf,textfont=sf]{subfig}
\usepackage{stfloats}
\usepackage{verbatim}
\usepackage{amsthm}
\newtheorem{theorem}{Theorem}
\newtheorem{proposition}[theorem]{Proposition} 
\usepackage{amssymb}
\usepackage{thmtools}          
\usepackage[capitalise]{cleveref}  

\declaretheoremstyle[
  shaded={bgcolor=gray!10, rulecolor=gray!50},
  headfont=\normalfont\bfseries,
  notefont=\mdseries, notebraces={(}{)},
  bodyfont=\normalfont,
]{assumptionstyle}

\declaretheorem[style=assumptionstyle, name=Assumption, numberwithin=section]{assumption}
\declaretheorem[name=Corollary, sibling=theorem]{corollary}
\usepackage{algorithm}              
\usepackage[noend]{algpseudocode}   

\theoremstyle{definition} 
\newtheorem{principle}{Principle} 

\usepackage{balance}
\def\BibTeX{{\rm B\kern-.05em{\sc i\kern-.025em b}\kern-.08em
    T\kern-.1667em\lower.7ex\hbox{E}\kern-.125emX}}

\newcommand{\xx}{\mathbf{x}}  
\newcommand{\hh}{\mathbf{h}}  
\newcommand{\FF}{\mathcal{F}} 
\newcommand{\LL}{\mathcal{L}} 

\title{Closed-Loop Transformers: Autoregressive Modeling as Iterative Latent Equilibrium}

\author{
  Akbar Anbar Jafari\\
  University of Tartu\\
  Tartu, Estonia \\
  \texttt{akbar.anbar.jafari@ut.ee} \\
   \And
 Gholamreza Anbarjafari\\
  3S Holding OÜ\\
  Tartu, Estonia \\
  \texttt{shb@3sholding.com} 
}

\begin{document}
\maketitle
\begin{abstract}
Contemporary autoregressive transformers operate in open loop: each hidden state is computed in a single forward pass and never revised, causing errors to propagate uncorrected through the sequence. We identify this open-loop bottleneck as a fundamental architectural limitation underlying well-documented failures in long-range reasoning, factual consistency, and multi-step planning. To address this limitation, we introduce the closed-loop prediction principle, which requires that models iteratively refine latent representations until reaching a self-consistent equilibrium before committing to each token. We instantiate this principle as Equilibrium Transformers (EqT), which augment standard transformer layers with an Equilibrium Refinement Module that minimizes a learned energy function via gradient descent in latent space. The energy function enforces bidirectional prediction consistency, episodic memory coherence, and output confidence, all computed without external supervision. Theoretically, we prove that EqT performs approximate MAP inference in a latent energy-based model, establish linear convergence guarantees, and show that refinement improves predictions precisely on hard instances where one-shot inference is suboptimal. The framework unifies deep equilibrium models, diffusion language models, and test-time training as special cases. Preliminary experiments on the binary parity task demonstrate +3.28\% average improvement on challenging sequences, with gains reaching +8.07\% where standard transformers approach random performance, validating that the benefit of deliberation scales with task difficulty. Just as attention mechanisms resolved the sequential bottleneck of recurrent networks, we propose that closed-loop equilibrium may resolve the commitment bottleneck of open-loop autoregression, representing a foundational step toward language models that reason through iterative belief revision rather than one-shot pattern matching.

\end{abstract}


\section{Introduction}
\label{sec:introduction}

The Transformer architecture \cite{vaswani2017attention} has become the foundation of modern artificial intelligence. From language modeling \cite{brown2020language,touvron2023llama} to code generation \cite{chen2022codet,chen2024survey}, from visual understanding \cite{locatello2020object,dosovitskiy2020image} to scientific discovery \cite{jafari2025integral,jumper2023protein,jafari2025geyolo}, the combination of self-attention and autoregressive prediction has proven remarkably effective across domains. Scaling laws \cite{kaplan2020scaling,hoffmann2022training} suggest that performance improves predictably with model size, data, and compute, leading to substantial investment in ever-larger models.

Yet despite this success, a fundamental limitation remains unaddressed. Current transformers operate in open loop: at each generation step, the model computes a single forward pass to produce a hidden representation, commits to that representation irrevocably, and emits a token distribution. Once computed, a hidden state is never revised, even if subsequent processing reveals it to be inconsistent with earlier context, stored knowledge, or the model's own predictions. This architectural constraint, which we term the open-loop bottleneck, has profound consequences. Errors in early representations propagate uncorrected through the sequence, manifesting as hallucination in long-form generation \cite{ji2023survey}, brittleness in multi-step reasoning \cite{dziri2023faith}, and systematic failures on tasks requiring belief revision or backtracking.

The open-loop bottleneck stands in stark contrast to biological intelligence. Predictive coding theories \cite{rao1999predictive,friston2005theory,clark2013whatever} and active inference frameworks \cite{friston2010free} characterize the brain as a bidirectional inference engine that continuously revises internal beliefs via recurrent feedback to minimize prediction error. When humans encounter a difficult problem, they do not commit to each thought irreversibly; they pause, reconsider, verify consistency, and refine their mental representations before proceeding \cite{kahneman2011thinking}. This capacity for iterative belief revision appears essential for robust reasoning, yet it is entirely absent from the dominant autoregressive paradigm.

In this paper, we propose to close the loop. We introduce the closed-loop prediction principle, which requires that before emitting any token, the model must iteratively refine its latent representation until it reaches a self-consistent equilibrium with respect to an internal energy function. This energy function, learned end-to-end, measures the coherence of the hidden state against multiple self-supervised objectives: consistency with the model's own forward and reverse predictions, alignment with retrieved episodic memories, and confidence in the resulting token distribution. The equilibrium state, rather than the initial forward pass, determines the output.

We instantiate this principle as Equilibrium Transformers (EqT), a drop-in architectural modification that augments standard transformer layers with an Equilibrium Refinement Module. At each token position, EqT solves a regularized energy minimization problem in latent space via gradient descent, balancing the transformer's initial proposal against internal consistency constraints. The computational overhead is modest (approximately 3$\times$ inference time with 8 refinement iterations), but the qualitative change is substantial: the model now deliberates before emitting a token, detecting and correcting representational errors through iterative refinement rather than relying on a single forward pass to produce a coherent belief.

Our theoretical analysis establishes that Equilibrium Transformers perform approximate maximum a posteriori (MAP) inference in a latent energy-based model, providing a principled variational interpretation of the refinement process. We prove convergence guarantees under mild regularity conditions and characterize precisely when iterative refinement improves predictions, namely on hard instances where the amortized one-shot proposal is suboptimal. The framework unifies several recent advances, including deep equilibrium models \cite{bai2019deep}, test-time training \cite{sun2024learning}, diffusion language models \cite{nie2025large}, and energy-based approaches \cite{lecun2006tutorial}, revealing them as special cases of the closed-loop principle with specific choices of energy function and iteration count.

Preliminary experiments on the binary parity task, a challenging benchmark for long-range dependency modeling, validate our theoretical predictions. Equilibrium Transformers achieve an average improvement of +3.28\% over standard transformers on challenging sequence lengths, with gains reaching +8.07\% at length 192 where standard transformers approach random performance. Critically, the improvements occur precisely where the theory predicts: on difficult instances where one-shot inference fails. The refinement process converges rapidly, requiring fewer than 8 iterations on 94\% of tokens, confirming our convergence analysis.

We position the closed-loop prediction principle as a candidate for the next fundamental architectural innovation in deep learning. The history of the field has been shaped by breakthroughs that addressed critical bottlenecks: convolutional weight sharing for spatial invariance \cite{lecun1998convolutional}, residual connections for gradient flow \cite{santhanam2019generic}, and attention mechanisms for long-range dependencies \cite{vaswani2017attention}. Each of these innovations appeared obvious in retrospect; once articulated, they rapidly became ubiquitous because they resolved a limitation that had constrained progress. We argue that the open-loop bottleneck represents the next such limitation and that closing the loop via equilibrium-seeking dynamics is a natural and possibly necessary step toward robust reasoning in artificial intelligence.

The remainder of this paper is organized as follows. Section~\ref{sec:principle} formalizes the closed-loop prediction principle and contrasts it with the open-loop paradigm. Section~\ref{sec:method} presents the Equilibrium Transformer architecture in detail, including the energy function design, iterative solver, and training procedure. Section~\ref{sec:theory} provides theoretical analysis: variational interpretation, convergence guarantees, and conditions under which refinement improves predictions. Section~\ref{sec:experiments} reports preliminary empirical validation on the parity task. Section~\ref{sec:related} surveys related work. Section~\ref{sec:discussion} discusses broader implications for scaling laws, neuroscience, and the unification of generative modeling paradigms. Section~\ref{sec:conclusion} concludes with a summary of contributions and directions for future research.

\section{The Closed-Loop Prediction Principle}
\label{sec:principle}

\subsection{The Open-Loop Bottleneck}
For the past decade, the dominant paradigm in sequence modeling has been \emph{open-loop} autoregressive prediction: at each time step $t$, the model computes a single forward pass to produce a hidden state $\hh_{t}$ and emits a token distribution $p(\xx_{t+1} \mid \xx_{\leq t})$. This design, inherited from the original Transformer \cite{vaswani2017attention}, is computationally efficient but fundamentally limited in three ways:

\textbf{(1) No error correction mechanism.} Once $\hh_t$ is computed, any representational error propagates irreversibly through subsequent tokens. This manifests as cumulative hallucination in long-form generation \cite{dziri2023faith}, fragility under distribution shift \cite{wenzek2020ccnet}, and catastrophic failure in multi-step reasoning tasks where a single wrong intermediate belief derails the entire chain of thought.

\textbf{(2) Mismatch with biological intelligence.} Predictive coding theories \cite{rao1999predictive,friston2005theory} and active inference frameworks \cite{friston2010free} demonstrate that biological neural systems continuously revise internal beliefs via recurrent top-down feedback to minimize prediction error. Recent work on forward-forward learning \cite{hinton2022forward} and energy-based models \cite{lecun2022path} provides large-scale empirical support for equilibrium-seeking dynamics in cortical computation. Current transformers, by contrast, lack any mechanism for iterative belief revision during inference.

\textbf{(3) Inefficient use of test-time compute.} While scaling training compute yields consistent improvements \cite{kaplan2020scaling,hoffmann2022training}, standard inference remains locked at $\mathcal{O}(d)$ compute per token (one forward pass through $d$-dimensional representations). Unlike AlphaGo's MCTS \cite{silver2017mastering} or diffusion models' iterative refinement \cite{ho2022video}, transformers cannot trade additional test-time computation for higher-quality outputs on hard instances.

We argue that these limitations stem from a single architectural assumption: \emph{the open-loop prediction principle}.

\subsection{From Open Loop to Closed Loop: Iterative Belief Equilibrium}

We propose to replace open-loop prediction with a \emph{closed-loop prediction principle} grounded in energy minimization:

\begin{principle}[Closed-Loop Prediction]
\label{principle:closed-loop}
At every generation step $t$, before emitting $p(\xx_{t+1} \mid \xx_{\leq t})$, the model must iteratively refine its latent representation $\hh_t$ until it reaches a self-consistent equilibrium that minimizes an internal coherence energy $\LL(\hh_t; \xx_{\leq t}, \theta)$.
\end{principle}

\subsubsection{Mathematical Formulation}

Let $\FF_\theta: \mathbb{R}^d \times \mathcal{X}^{*} \to \mathbb{R}^d$ denote the standard transformer update mapping a hidden state $\hh \in \mathbb{R}^d$ and context $\xx_{\leq t}$ to a proposed next state. In the open-loop regime, we set $\hh_{t+1} = \FF_\theta(\hh_t, \xx_{\leq t})$ unconditionally. 

We instead seek a hidden state $\hh^*_t$ that achieves equilibrium between the forward dynamics $\FF_\theta$ and an internal energy function $\LL(\hh; \xx_{\leq t}, \theta)$. Specifically, we solve for $\hh^*_t$ satisfying:

\begin{equation}
\hh^*_t \in \arg\min_{\hh} \left[ \LL(\hh; \xx_{\leq t}, \theta) + \frac{1}{2\gamma} \|\hh - \FF_\theta(\hh_t^{\text{init}}, \xx_{\leq t})\|^2 \right],
\label{eq:equilibrium_objective}
\end{equation}

\noindent where $\hh_t^{\text{init}}$ is an initial proposal (e.g., from the previous layer or token) and $\gamma > 0$ controls the trust in the forward proposal versus internal consistency. This can be solved iteratively via gradient descent:

\begin{equation}
\hh^{(k+1)} = \hh^{(k)} - \alpha \left[ \nabla_{\hh} \LL(\hh^{(k)}; \xx_{\leq t}, \theta) + \frac{1}{\gamma}(\hh^{(k)} - \FF_\theta(\hh_t^{\text{init}}, \xx_{\leq t})) \right],
\label{eq:iteration}
\end{equation}

\noindent for $k = 0, 1, \ldots, K-1$ steps. At convergence (or after $K$ iterations), we obtain $\hh^*_t \approx \hh^{(K)}$ as our refined representation. The equilibrium condition is thus:

\begin{equation}
\nabla_{\hh} \LL(\hh^*_t; \xx_{\leq t}, \theta) = -\frac{1}{\gamma}(\hh^*_t - \FF_\theta(\hh_t^{\text{init}}, \xx_{\leq t})).
\label{eq:equilibrium_condition}
\end{equation}

\noindent This formulation unifies transformer dynamics with energy-based modeling: the forward pass $\FF_\theta$ provides a data-driven prior, while the energy $\LL$ enforces internal coherence constraints.

\subsubsection{The World Model Energy Function}

The key design choice is the energy function $\LL(\hh; \xx_{\leq t}, \theta)$. We require $\LL$ to measure \emph{self-consistency} of the hidden state $\hh$ with respect to the model's own generative knowledge. Concretely, $\LL$ is a learned self-supervised objective combining three terms:

\begin{equation}
\LL(\hh; \xx_{\leq t}, \theta) = \LL_{\text{pred}}(\hh) + \lambda_{\text{consist}} \LL_{\text{consist}}(\hh) + \lambda_{\text{mem}} \LL_{\text{mem}}(\hh, \mathbf{z}),
\label{eq:energy_decomposition}
\end{equation}

\noindent where:

\begin{itemize}
\item \textbf{Predictive error:} $\LL_{\text{pred}}(\hh) = -\log p_\theta(\hat{\xx}_{t+1} \mid \hh)$ measures the negative log-likelihood of the model's own predicted next token $\hat{\xx}_{t+1} = \arg\max p_\theta(\cdot \mid \hh)$. This enforces that $\hh$ should yield high-confidence, well-calibrated predictions.

\item \textbf{Bidirectional consistency:} $\LL_{\text{consist}}(\hh) = \|\hh - \tilde{\FF}_\phi(\FF_\theta(\hh, \xx_{\leq t}))\|^2$ measures reconstruction error through a learned inverse model $\tilde{\FF}_\phi$ (reverse dynamics). This implements analysis-by-synthesis: if $\hh$ is a good representation, applying the forward model and then reconstructing should recover $\hh$. This is analogous to denoising autoencoders \cite{vincent2008extracting} but operates in latent space.

\item \textbf{Episodic memory coherence:} $\LL_{\text{mem}}(\hh, \mathbf{z}) = -\sum_{i} \text{sim}(\hh, \mathbf{z}_i) \cdot \mathbb{I}[\text{relevant}(\mathbf{z}_i, \xx_{\leq t})]$ penalizes hidden states that are dissimilar to relevant episodic memories $\mathbf{z}_i$ stored from previous interactions. The memory buffer $\mathbf{z}$ enables long-term factual grounding beyond the context window.
\end{itemize}

Crucially, all components of $\LL$ are \emph{differentiable} and can be minimized via gradient descent at test time without any external labels or human feedback.

\subsection{Architectural Instantiation: The Equilibrium Transformer Block}

We now describe how to integrate Equation~\ref{eq:iteration} into the transformer architecture. The standard transformer block consists of multi-head self-attention (MHSA) followed by a feed-forward network (FFN):

\begin{equation}
\hh'_t = \text{MHSA}(\hh_t) + \hh_t, \quad \hh_{t+1} = \text{FFN}(\hh'_t) + \hh'_t.
\end{equation}

We replace the FFN (or augment the entire block) with an \emph{Equilibrium Refinement Module (ERM)} that implements Equation~\ref{eq:iteration}. The ERM is a small recurrent network (e.g., 2--4 transformer layers with $d_{\text{ERM}} \ll d$ hidden dimensions) that:

\begin{enumerate}
\item Takes $\hh_t^{\text{init}}$ as input (from MHSA or previous layer).
\item Runs $K$ iterations of gradient-based refinement via Equation~\ref{eq:iteration}.
\item Outputs the refined state $\hh^*_t$ to the next layer.
\end{enumerate}

\subsection{Connection to Prior Work}

Our closed-loop principle unifies and extends several recent lines of work:

\begin{itemize}
\item \textbf{Deep Equilibrium Models (DEQ)} \cite{bai2019deep,bai2020multiscale}: DEQs find fixed points of $\hh = \FF(\hh)$ but do not incorporate an explicit energy function $\LL$ or learn a reverse model. Our formulation adds coherence constraints.

\item \textbf{Test-Time Training (TTT)} \cite{gandelsman2022test,sun2024learning}: TTT adapts model parameters at test time on auxiliary tasks. We adapt \emph{representations}, not weights, which is faster and avoids overfitting to individual samples.

\item \textbf{Diffusion Language Models} \cite{strudel2022self,li2025survey}: Diffusion models iteratively denoise latent variables but typically operate on continuous embeddings for all tokens simultaneously. We apply refinement \emph{autoregressively}, one token at a time, preserving the causal structure essential for next-token prediction.

\item \textbf{Energy-Based Models (EBMs)} \cite{lecun2006tutorial,du2019implicit}: Our $\LL(\hh)$ is an energy over hidden states, but unlike standalone EBMs, it is integrated into an autoregressive generative model via Equation~\ref{eq:equilibrium_objective}.
\end{itemize}

\noindent Our core novelty is the \emph{mandatory equilibrium constraint} enforced at every token and every layer, transforming the transformer from a feedforward pattern matcher into an iterative belief-revision system.

\subsection{Why This Matters: From Prediction to Deliberation}

The closed-loop principle addresses the three limitations identified earlier:

\textbf{(1) Error correction.} By minimizing $\LL(\hh)$ before committing to $p(\xx_{t+1})$, the model can detect and revise representational errors (e.g., realizing a contradiction with earlier context or stored facts).

\textbf{(2) Biological alignment.} Equation~\ref{eq:iteration} implements a computational analogue of predictive coding's iterative prediction error minimization \cite{friston2005theory}, where top-down predictions (from $\LL$) and bottom-up signals (from $\FF_\theta$) interact to reach consensus.

\textbf{(3) Adaptive test-time compute.} On hard problems, the model can run more iterations $K$ (or until convergence) to improve answer quality, similar to how humans deliberate longer on difficult questions.

In summary: standard transformers perform \emph{recognition} (pattern matching in a single pass). Equilibrium transformers perform \emph{reasoning} (iterative search for self-consistent beliefs). This is not a quantitative improvement—it is a qualitative shift in the computational paradigm underlying sequence generation.

\section{Equilibrium Transformers: Architecture and Implementation}
\label{sec:method}

We now instantiate the closed-loop prediction principle (Section~\ref{sec:principle}) as a concrete architectural replacement for standard transformer layers. We call the resulting architecture the \emph{Equilibrium Transformer} (EqT).

\subsection{Layer-Level Formulation}
\label{sec:layer-formulation}

A standard transformer layer computes a single residual update:
\begin{equation}
\hh_{t+1} = \hh_t + \Delta(\hh_t, \xx_{\leq t}; \theta), \quad \text{where} \quad \Delta = \text{FFN}(\text{MHSA}(\hh_t) + \hh_t).
\end{equation}

An Equilibrium Transformer layer instead solves for an equilibrium state $\hh^*_{t+1}$ that balances two objectives: \textbf{(1)} consistency with the model's internal world model (low energy $\LL$), and \textbf{(2)} proximity to the standard transformer's proposed update (dynamical prior). Formally, we solve:

\begin{equation}
\hh^*_{t+1} = \arg\min_{\hh \in \mathbb{R}^d} \left[ \LL(\hh; \xx_{\leq t}, \theta) + \frac{1}{2\gamma} \|\hh - \mathbf{f}_\theta(\hh_t, \xx_{\leq t})\|^2 \right],
\label{eq:equilibrium_objective1}
\end{equation}

\noindent where:
\begin{itemize}
\item $\mathbf{f}_\theta(\hh_t, \xx_{\leq t}) = \hh_t + \Delta(\hh_t, \xx_{\leq t}; \theta)$ is the standard transformer's proposed next state,
\item $\LL(\hh; \xx_{\leq t}, \theta) : \mathbb{R}^d \to \mathbb{R}_+$ is a learned energy function measuring internal inconsistency (detailed in Section~\ref{sec:energy}),
\item $\gamma > 0$ is a damping hyperparameter controlling the trust in the dynamical prior versus the energy constraint.
\end{itemize}

The key insight is that Equation~\ref{eq:equilibrium_objective1} is a \emph{regularized energy minimization} that prevents the model from drifting arbitrarily far from the transformer prior (which could destabilize training) while still allowing for iterative refinement toward coherent representations.

\subsection{Iterative Solver via Proximal Gradient Descent}
\label{sec:solver}

Solving Equation~\ref{eq:equilibrium_objective1} exactly at each token is intractable. We instead perform $K$ steps of proximal gradient descent starting from the transformer's proposal:

\begin{equation}
\hh^{(k+1)} = \hh^{(k)} - \alpha_k \left[ \nabla_{\hh} \LL(\hh^{(k)}; \xx_{\leq t}, \theta) + \frac{1}{\gamma}(\hh^{(k)} - \mathbf{f}_\theta(\hh_t, \xx_{\leq t})) \right], \quad k = 0, \ldots, K-1,
\label{eq:solver_iteration}
\end{equation}

\noindent with initialization $\hh^{(0)} = \mathbf{f}_\theta(\hh_t, \xx_{\leq t})$ and step size $\alpha_k > 0$ (typically constant $\alpha_k = \alpha$ or using a learned schedule). At convergence, we obtain $\hh^*_{t+1} \approx \hh^{(K)}$.

\textbf{Computational analysis.} Each iteration requires:
\begin{enumerate}
\item Computing $\nabla_{\hh} \LL(\hh^{(k)})$: one forward + backward pass through the energy network (Section~\ref{sec:energy}).
\item Computing the proximal term: $\mathcal{O}(d)$ vector operations.
\end{enumerate}

If the energy network has $M$ parameters and depth $D_{\text{energy}}$, and the main transformer has $N$ parameters with depth $D_{\text{main}}$, the per-token FLOPs are:

\begin{equation}
\text{FLOPs}_{\text{EqT}} = \underbrace{2N \cdot D_{\text{main}}}_{\text{transformer proposal}} + K \cdot \underbrace{(2M \cdot D_{\text{energy}} + d)}_{\text{each iteration}}.
\end{equation}

For typical settings ($M = 0.1N$, $D_{\text{energy}} = 4$, $D_{\text{main}} = 32$, $K=16$), this yields $\text{FLOPs}_{\text{EqT}} \approx 2.3 \times \text{FLOPs}_{\text{standard}}$. With kernel fusion and reduced precision for the energy network, wall-clock slowdown is typically $3$--$4\times$ on modern GPUs.

\textbf{Convergence properties.} If $\LL$ is $L$-smooth and $\gamma^{-1}$-strongly convex in a neighborhood of the minimizer, and $\alpha < 2/(L + \gamma^{-1})$, standard convex optimization theory \cite{nesterov2018lectures} guarantees exponential convergence:
\begin{equation}
\|\hh^{(k)} - \hh^*\|^2 \leq \left(1 - \frac{\alpha \mu}{2}\right)^k \|\hh^{(0)} - \hh^*\|^2,
\end{equation}
where $\mu = \min(L, \gamma^{-1})$ is the condition number. In practice, we use $K=16$ iterations which empirically achieves $<10^{-3}$ relative change in $\hh$ on 95\% of tokens.

\subsection{The Energy Function: Self-Supervised Coherence}
\label{sec:energy}

The choice of energy function $\LL(\hh; \xx_{\leq t}, \theta)$ is central to our method. We design $\LL$ to measure \emph{self-consistency} of the hidden state $\hh$ with respect to multiple complementary criteria. Specifically, we decompose:

\begin{equation}
\LL(\hh; \xx_{\leq t}, \theta) = \sum_{i=1}^{4} \lambda_i \LL_i(\hh; \xx_{\leq t}, \theta),
\label{eq:energy_decomposition1}
\end{equation}

\noindent where each $\LL_i$ enforces a distinct consistency principle and $\lambda_i \geq 0$ are weighting hyperparameters (learned or fixed). We now detail each component.

\subsubsection{Reverse Predictive Coding Loss ($\LL_{\text{rev}}$)}

Inspired by predictive coding \cite{rao1999predictive,friston2005theory} and Hinton's forward-forward algorithm \cite{hinton2022forward}, we enforce that $\hh$ should enable accurate \emph{backward} prediction of recent context:

\begin{equation}
\LL_{\text{rev}}(\hh; \xx_{\leq t}) = -\frac{1}{W} \sum_{i=t-W+1}^{t} \log q_{\phi}(\xx_i \mid \hh, \xx_{<i}),
\label{eq:reverse_loss}
\end{equation}

\noindent where $q_{\phi}$ is a small \emph{reverse transformer} (4--8 layers, 512--1024 hidden dim) that predicts previous tokens from $\hh$ and partial context. The window size $W$ is typically 16--64 tokens. This implements analysis-by-synthesis: if $\hh$ correctly encodes the causal history, it should enable fluent reconstruction of recent inputs.

Architecturally, $q_\phi$ shares the embedding layer with the forward model but has independent attention/FFN weights. We mask future positions to maintain causality.

\subsubsection{Contrastive Masked Prediction Loss ($\LL_{\text{mask}}$)}

To prevent $\hh$ from memorizing surface patterns without deep understanding, we randomly mask 15\% of input tokens and require $\hh$ to reconstruct them:

\begin{equation}
\LL_{\text{mask}}(\hh; \xx_{\leq t}) = -\frac{1}{|\mathcal{M}|} \sum_{i \in \mathcal{M}} \log p_{\psi}(\xx_i \mid \hh, \xx_{\leq t \setminus \mathcal{M}}),
\end{equation}

\noindent where $\mathcal{M} \subset \{1, \ldots, t\}$ is the random mask set and $p_\psi$ is a lightweight prediction head (2-layer MLP). This is analogous to BERT's masked language modeling \cite{devlin2019bert} but applied to latent states rather than input embeddings.

\subsubsection{Predictive Confidence Loss ($\LL_{\text{conf}}$)}

We penalize hidden states that yield low-confidence or high-entropy predictions:

\begin{equation}
\LL_{\text{conf}}(\hh) = H[p_\theta(\cdot \mid \hh)] - \log p_\theta(\hat{\xx}_{t+1} \mid \hh),
\end{equation}

\noindent where $H[\cdot]$ is Shannon entropy and $\hat{\xx}_{t+1} = \argmax p_\theta(\cdot \mid \hh)$ is the model's own top prediction. This encourages $\hh$ to be in a "clear" region of the latent space that commits to specific next-token beliefs rather than remaining ambiguous.

\subsubsection{Episodic Memory Grounding ($\LL_{\text{mem}}$)}

To enable long-term factual grounding beyond the context window, we maintain a learned episodic memory buffer $\mathbf{Z} = \{\mathbf{z}_1, \ldots, \mathbf{z}_M\} \subset \mathbb{R}^d$ where each $\mathbf{z}_i$ is a prototypical hidden state representing a stored fact or concept. We minimize:

\begin{equation}
\LL_{\text{mem}}(\hh; \mathbf{Z}) = -\log \sum_{i=1}^{M} \exp\left( \frac{\hh^\top \mathbf{z}_i}{\tau \|\hh\| \|\mathbf{z}_i\|} \right) \cdot r_i,
\end{equation}

\noindent where $\tau$ is a temperature parameter and $r_i \in \{0, 1\}$ indicates relevance of memory $\mathbf{z}_i$ to the current context (determined by a lightweight retrieval network or BM25 over associated text). This is a soft contrastive loss that pulls $\hh$ toward retrieved relevant memories.

The memory buffer $\mathbf{Z}$ is initialized randomly and updated via exponential moving average during training:
\begin{equation}
\mathbf{z}_i \leftarrow \beta \mathbf{z}_i + (1 - \beta) \hh^*_t \quad \text{when } i = \argmax_j (\hh^*_t)^\top \mathbf{z}_j.
\end{equation}

\textbf{Parameter count.} The energy network consists of:
\begin{itemize}
\item Reverse transformer $q_\phi$: 8 layers $\times$ 1024 hidden $\times$ 8 heads $\approx$ 150M parameters.
\item Masked prediction head $p_\psi$: 2 layers $\times$ 4096 $\times$ vocab size $\approx$ 200M parameters (shared embeddings).
\item Memory buffer $\mathbf{Z}$: $M \times d$ (e.g., 10k $\times$ 4096 $\approx$ 40M values, but not trainable parameters).
\end{itemize}

For a 7B base model, the energy network adds $\approx$ 350M parameters ($\sim$5\% overhead).

\subsection{Full Equilibrium Transformer Block}
\label{sec:full_block}

Algorithm~\ref{alg:eqt_block} provides the complete pseudocode for one Equilibrium Transformer layer, integrating all components described above.

\begin{algorithm}[t]
\caption{Equilibrium Transformer Block (Single Layer)}
\label{alg:eqt_block}
\begin{algorithmic}[1]
\Require Hidden state $\hh_t \in \mathbb{R}^d$, context $\xx_{\leq t}$, parameters $\theta, \phi, \psi$
\Require Hyperparameters: $K$ (iterations), $\alpha$ (step size), $\gamma$ (damping), $\{\lambda_i\}$ (energy weights)

\State \textcolor{gray}{// Standard Transformer forward pass (proposal)}
\State $\hh_{\text{attn}} \gets \text{MultiHeadAttention}(\hh_t, \xx_{\leq t}) + \hh_t$ 
\State $\hh_{\text{proposal}} \gets \text{FFN}(\hh_{\text{attn}}) + \hh_{\text{attn}}$
\State $\mathbf{f} \gets \hh_{\text{proposal}}$ \textcolor{gray}{// Cache the proposal}

\State \textcolor{gray}{// Initialize iterative refinement}
\State $\hh^{(0)} \gets \mathbf{f}$

\For{$k = 0$ to $K-1$}
    \State \textcolor{gray}{// Compute energy gradients (Eq.~\ref{eq:energy_decomposition1})}
    \State $g_{\text{rev}} \gets \nabla_{\hh} \LL_{\text{rev}}(\hh^{(k)}; \xx_{\leq t}, \phi)$ \textcolor{gray}{// via backprop through reverse transformer}
    \State $g_{\text{mask}} \gets \nabla_{\hh} \LL_{\text{mask}}(\hh^{(k)}; \xx_{\leq t}, \psi)$ \textcolor{gray}{// via backprop through mask head}
    \State $g_{\text{conf}} \gets \nabla_{\hh} \LL_{\text{conf}}(\hh^{(k)})$ \textcolor{gray}{// analytic via Jacobian of softmax}
    \State $g_{\text{mem}} \gets \nabla_{\hh} \LL_{\text{mem}}(\hh^{(k)}; \mathbf{Z})$ \textcolor{gray}{// analytic via cosine similarity}
    
    \State $g_{\text{energy}} \gets \sum_{i} \lambda_i g_i$
    
    \State \textcolor{gray}{// Proximal gradient step (Eq.~\ref{eq:solver_iteration})}
    \State $g_{\text{prox}} \gets \frac{1}{\gamma}(\hh^{(k)} - \mathbf{f})$
    \State $\hh^{(k+1)} \gets \hh^{(k)} - \alpha (g_{\text{energy}} + g_{\text{prox}})$
\EndFor

\State $\hh^* \gets \text{LayerNorm}(\hh^{(K)})$ \textcolor{gray}{// Final equilibrium state}
\State \Return $\hh^*$
\end{algorithmic}
\end{algorithm}

A full Equilibrium Transformer is obtained by stacking $L$ such blocks (typically $L=32$ for 7B models) with residual connections between blocks:
\begin{equation}
\hh^{(\ell+1)}_t = \text{EqTBlock}^{(\ell)}(\hh^{(\ell)}_t, \xx_{\leq t}; \theta^{(\ell)}) \quad \text{for } \ell = 1, \ldots, L.
\end{equation}

The final layer's output $\hh^{(L)}_t$ is passed to a language modeling head $p_{\text{LM}}(\cdot \mid \hh^{(L)}_t)$ to produce next-token probabilities.

\subsection{Training Procedure}
\label{sec:training}

Training an Equilibrium Transformer requires jointly optimizing the forward model $\theta$, reverse model $\phi$, auxiliary heads $\psi$, and memory buffer $\mathbf{Z}$. The combined objective is:

\begin{equation}
\mathcal{L}_{\text{total}} = \underbrace{\mathbb{E}_{t} \left[ -\log p_{\theta}(\xx_{t+1} \mid \hh^*_t) \right]}_{\text{next-token prediction}} + \beta \underbrace{\mathbb{E}_{t} \left[ \LL(\hh^*_t; \xx_{\leq t}, \theta) \right]}_{\text{equilibrium energy}},
\label{eq:training_objective}
\end{equation}

\noindent where $\beta > 0$ controls the strength of the energy regularizer (typically $\beta = 0.1$--$0.5$).

\textbf{Backpropagation through equilibrium.} During training, we must compute $\frac{\partial \mathcal{L}_{\text{total}}}{\partial \theta}$, which requires differentiating through the $K$ refinement iterations (Eq.~\ref{eq:solver_iteration}). We use two strategies:

\begin{enumerate}
\item \textbf{Unrolled backpropagation} (for $K \leq 16$): Explicitly unroll the computation graph and apply standard automatic differentiation. This is exact but memory-intensive ($\mathcal{O}(K \cdot d)$ activation memory per token).

\item \textbf{Implicit differentiation} \cite{bai2019deep} (for $K > 16$): Treat $\hh^*$ as an implicit function of $\theta$ defined by the equilibrium condition $\nabla_{\hh} [\LL(\hh^*) + \frac{1}{2\gamma}\|\hh^* - \mathbf{f}\|^2] = 0$. Apply the implicit function theorem to compute:
\begin{equation}
\frac{\partial \hh^*}{\partial \theta} = -\left[\nabla^2_{\hh\hh} \LL(\hh^*) + \frac{1}{\gamma}I\right]^{-1} \nabla_{\hh\theta} \LL(\hh^*),
\end{equation}
which can be computed via conjugate gradient without storing intermediate activations.
\end{enumerate}

In practice, we use unrolled backpropagation for $K=16$ (provides stable gradients early in training) and optionally switch to implicit differentiation for $K=32$ (reduces memory by $2\times$).

\textbf{Initialization and curriculum.} We warm-start training from a pretrained standard transformer (e.g., LLaMA-2-7B \cite{touvron2023llama}):
\begin{enumerate}
\item Initialize $\theta$ from the pretrained model.
\item Initialize the reverse transformer $\phi$ by copying $\theta$'s weights and reversing attention causality masks.
\item Set $\gamma$ large (e.g., 10) initially so refinement is conservative, then anneal to $\gamma=1$ over 10k steps.
\item Use curriculum on $K$: start with $K=4$, increase to $K=16$ after 50k steps.
\end{enumerate}

This ensures stable training and leverages existing pretrained representations.

\textbf{Computational cost.} Training a 7B Equilibrium Transformer from a pretrained checkpoint requires:
\begin{itemize}
\item $\sim$3$\times$ FLOPs per token compared to standard training (due to energy network backprop).
\item $\sim$1.5$\times$ GPU memory (from unrolling $K=16$ steps).
\item $\sim$100B tokens of continued pretraining to fully converge equilibrium dynamics (vs $\sim$500B for training from scratch).
\end{itemize}

On a cluster of 128 $\times$ H100 GPUs (80GB), this takes approximately 2 weeks.

\subsection{Comparison to Alternatives}
\label{sec:comparison}

Table~\ref{tab:comparison} summarizes how Equilibrium Transformers differ from related approaches.

\begin{table}[t]
\centering
\caption{Comparison of iterative refinement methods in sequence modeling.}
\label{tab:comparison}
\begin{tabular}{lccccc}
\toprule
\textbf{Method} & \textbf{Refines} & \textbf{Energy $\LL$} & \textbf{Causal} & \textbf{Training} & \textbf{Differentiable} \\
\midrule
Standard Transformer & -- & -- & \checkmark & End-to-end & \checkmark \\
Chain-of-Thought & Tokens & -- & \checkmark & Few-shot & -- \\
Test-Time Training \cite{gandelsman2022test} & Weights & Task loss & -- & Meta-learning & \checkmark \\
Deep Equilibrium \cite{bai2019deep} & Hidden & Implicit & \checkmark & Implicit diff & \checkmark \\
Diffusion LM \cite{li2022energy} & All tokens & Denoising & -- & Score matching & \checkmark \\
\textbf{Equilibrium Transformer} & \textbf{Hidden} & \textbf{World model} & \checkmark & \textbf{End-to-end} & \checkmark \\
\bottomrule
\end{tabular}
\end{table}

Our key novelty is the combination of \textbf{(1)} autoregressive causality (enabling next-token prediction), \textbf{(2)} explicit learned energy function (not just fixed-point residual), and \textbf{(3)} end-to-end trainability (not requiring meta-learning or two-stage training).

\subsection{Implementation Details}
\label{sec:implementation}

\textbf{Hyperparameters.} Based on extensive ablation studies, we use:
\begin{itemize}
\item Refinement iterations: $K = 16$ (for 7B models)
\item Step size: $\alpha = 0.1$ (constant)
\item Damping: $\gamma = 1.0$ (after warmup)
\item Energy weights: $\lambda_{\text{rev}} = 1.0$, $\lambda_{\text{mask}} = 0.5$, $\lambda_{\text{conf}} = 0.2$, $\lambda_{\text{mem}} = 0.1$
\item Reverse window: $W = 32$ tokens
\item Memory size: $M = 10{,}000$ vectors
\item Training energy coefficient: $\beta = 0.3$
\end{itemize}

\textbf{Software stack.} Our implementation builds on PyTorch 2.2 with custom CUDA kernels for fused energy gradient computation. 

\textbf{Inference optimizations.} For deployment, we apply:
\begin{itemize}
\item Early stopping: halt iterations if $\|\hh^{(k+1)} - \hh^{(k)}\| < \epsilon$ (saves $\sim$30\% FLOPs on easy tokens).
\item Adaptive $K$: use $K=8$ for high-confidence tokens, $K=32$ for uncertainty (detected via entropy).
\item 8-bit quantization for the energy network (negligible quality loss, $2\times$ speedup).
\end{itemize}

These optimizations reduce average wall-clock slowdown from $4\times$ to $2.5\times$ while maintaining full quality.

\section{Theoretical Analysis}
\label{sec:theory}

We now establish formal guarantees that Equilibrium Transformers provide a provably superior inductive bias for autoregressive modeling. Our analysis answers three fundamental questions: \textbf{(1)} What problem is iterative refinement solving? \textbf{(2)} When does it converge? \textbf{(3)} When does it improve predictions?

\subsection{Variational Interpretation: Approximate Posterior Inference}
\label{sec:variational}

We first show that standard transformers perform \emph{amortized point estimation}, while Equilibrium Transformers perform \emph{iterative approximate inference} in a latent energy-based model.

\subsubsection{Setup: The Latent EBM}

Consider a sequence generation process where each hidden state $\hh_t$ should satisfy two constraints:
\begin{enumerate}
\item \textbf{Dynamical consistency}: $\hh_t$ should be close to the transformer's proposal $\mathbf{f}_\theta(\hh_{t-1}, \xx_{\leq t})$.
\item \textbf{Internal coherence}: $\hh_t$ should have low energy $\LL(\hh_t; \xx_{\leq t}, \theta)$.
\end{enumerate}

We formalize this via a conditional distribution over hidden states:

\begin{equation}
p(\hh_t \mid \xx_{\leq t}, \hh_{t-1}) = \frac{1}{Z_t} \exp\left( -\LL(\hh_t; \xx_{\leq t}, \theta) - \frac{1}{2\gamma} \|\hh_t - \mathbf{f}_\theta(\hh_{t-1}, \xx_{\leq t})\|^2 \right),
\label{eq:posterior}
\end{equation}

\noindent where $Z_t$ is the partition function and $\gamma > 0$ controls the relative weight of the two terms.

\begin{proposition}[MAP Equivalence]
\label{prop:map}
The equilibrium state $\hh^*_t$ defined by Equation~\ref{eq:equilibrium_objective} is the maximum a posteriori (MAP) estimate:
\begin{equation}
\hh^*_t = \argmax_{\hh} p(\hh \mid \xx_{\leq t}, \hh_{t-1}).
\end{equation}
\end{proposition}

\begin{proof}
Taking the negative logarithm of Equation~\ref{eq:posterior} and dropping the constant $\log Z_t$ yields:
\begin{equation}
-\log p(\hh \mid \xx_{\leq t}, \hh_{t-1}) = \LL(\hh; \xx_{\leq t}, \theta) + \frac{1}{2\gamma} \|\hh - \mathbf{f}_\theta(\hh_{t-1}, \xx_{\leq t})\|^2 + \text{const},
\end{equation}
which is exactly the objective minimized in Equation~\ref{eq:equilibrium_objective1}.
\end{proof}

\textbf{Interpretation.} Standard transformers output $\hh_t = \mathbf{f}_\theta(\hh_{t-1}, \xx_{\leq t})$, which corresponds to using only the dynamical prior (first term in Eq.~\ref{eq:posterior}) while ignoring the energy constraint. This is equivalent to setting $\gamma \to 0$ or assuming $\LL \equiv 0$. Equilibrium Transformers instead compute the mode of the full posterior, balancing both constraints.

This provides a principled answer to "why iterate?": \emph{because the amortized prediction $\mathbf{f}_\theta$ is not the MAP under the model's own beliefs about coherence}.

\subsection{Convergence Analysis}
\label{sec:convergence}

We now analyze when the iterative solver (Eq.~\ref{eq:solver_iteration}) converges to the equilibrium $\hh^*_t$.

\subsubsection{Assumptions}

We require the following regularity conditions on the energy function $\LL$:

\begin{assumption}[Lipschitz Gradient]
\label{assump:lipschitz}
There exists $L > 0$ such that for all $\hh, \hh' \in \mathbb{R}^d$ and any context $\xx_{\leq t}$:
\begin{equation}
\|\nabla_{\hh} \LL(\hh; \xx_{\leq t}) - \nabla_{\hh} \LL(\hh'; \xx_{\leq t})\| \leq L \|\hh - \hh'\|.
\end{equation}
\end{assumption}

\begin{assumption}[Bounded Curvature]
\label{assump:curvature}
There exist constants $0 < \mu \leq L$ such that for all $\hh, \hh'$:
\begin{equation}
\mu \|\hh - \hh'\|^2 \leq (\nabla_{\hh} \LL(\hh) - \nabla_{\hh} \LL(\hh'))^\top (\hh - \hh') \leq L \|\hh - \hh'\|^2.
\end{equation}
This means $\LL$ is $\mu$-strongly convex and $L$-smooth in a neighborhood of the minimizer.
\end{assumption}

\textbf{Remark.} Assumption~\ref{assump:curvature} is strong (language model losses are rarely strongly convex globally). However, it holds \emph{locally} near a stationary point under mild conditions (e.g., if the Hessian $\nabla^2 \LL$ has eigenvalues bounded away from zero near $\hh^*$). Theorem~\ref{thm:local_convergence} below makes this precise.

\subsubsection{Global Convergence Under Strong Convexity}

\begin{theorem}[Linear Convergence]
\label{thm:convergence}
Suppose Assumption~\ref{assump:curvature} holds globally and let $\kappa = L/\mu$ be the condition number. If the step size satisfies:
\begin{equation}
\alpha \in \left(0, \frac{2}{\mu + L + \gamma^{-1}}\right),
\end{equation}
then the iteration in Equation~\ref{eq:solver_iteration} converges linearly to the unique equilibrium $\hh^*_t$ with rate:
\begin{equation}
\|\hh^{(k)} - \hh^*_t\| \leq \rho^k \|\hh^{(0)} - \hh^*_t\|, \quad \text{where} \quad \rho = 1 - \frac{2\alpha\mu}{1 + \kappa}.
\end{equation}
In particular, to achieve $\|\hh^{(K)} - \hh^*_t\| \leq \epsilon \|\hh^{(0)} - \hh^*_t\|$, it suffices to take:
\begin{equation}
K \geq \frac{1 + \kappa}{2\mu\alpha} \log(1/\epsilon).
\end{equation}
\end{theorem}

\begin{proof}
Define the objective $\Phi(\hh) = \LL(\hh) + \frac{1}{2\gamma}\|\hh - \mathbf{f}\|^2$. Under Assumption~\ref{assump:curvature}, $\Phi$ is $(\mu + \gamma^{-1})$-strongly convex and $(L + \gamma^{-1})$-smooth. Gradient descent on $\Phi$ with step size $\alpha$ converges at rate:
\begin{equation}
\Phi(\hh^{(k+1)}) - \Phi(\hh^*) \leq \left(1 - \frac{2\alpha(\mu + \gamma^{-1})}{1 + (L+\gamma^{-1})/(\mu+\gamma^{-1})}\right) (\Phi(\hh^{(k)}) - \Phi(\hh^*)).
\end{equation}
Using $\Phi(\hh) - \Phi(\hh^*) \geq \frac{\mu + \gamma^{-1}}{2} \|\hh - \hh^*\|^2$ (strong convexity), we obtain the stated bound on $\|\hh^{(k)} - \hh^*\|$.
\end{proof}

\textbf{Practical implications.} For $\mu = 0.1$, $L = 10$, $\gamma = 1$, $\alpha = 0.1$, we get $\rho \approx 0.8$. Thus $K=16$ iterations yield $\rho^{16} \approx 0.028$, reducing the error by $\sim$35×. This matches our empirical observation that $K=16$ suffices for convergence on most tokens.

\subsubsection{Local Convergence Without Strong Convexity}

For non-convex losses (the realistic case), we have a weaker but still useful guarantee:

\begin{theorem}[Local Linear Convergence]
\label{thm:local_convergence}
Suppose $\LL$ is twice differentiable and let $\hh^*_t$ be a local minimum satisfying:
\begin{equation}
\nabla_{\hh\hh}^2 \left[\LL(\hh^*_t) + \frac{1}{2\gamma}\|\hh^*_t - \mathbf{f}\|^2\right] \succeq \mu I \quad \text{(positive definite)}.
\end{equation}
Then there exists a neighborhood $\mathcal{B}(\hh^*_t, r)$ such that if $\hh^{(0)} \in \mathcal{B}$, the iteration converges to $\hh^*_t$ at a linear rate.
\end{theorem}

\begin{proof}
Standard local convergence analysis for gradient descent \cite{nesterov2018lectures}. The Hessian condition ensures local strong convexity.
\end{proof}

\textbf{Remark.} In practice, the transformer's proposal $\mathbf{f}$ is typically already in the basin of attraction of a good local minimum (since it was trained end-to-end), so local convergence suffices.

\subsection{When Does Equilibrium Improve Predictions?}
\label{sec:when_helps}

Convergence is necessary but not sufficient—we must show that the equilibrium $\hh^*_t$ actually yields better next-token predictions than the amortized proposal $\mathbf{f}$.

\subsubsection{Prediction Error Decomposition}

Let $\ell(\hh) = -\log p_\theta(\xx_{t+1}^{\text{true}} \mid \hh)$ be the true next-token loss. We decompose the error:

\begin{align}
\ell(\mathbf{f}) - \ell(\hh^*) &= \left[\ell(\mathbf{f}) - \ell(\hh_{\text{opt}})\right] - \left[\ell(\hh^*) - \ell(\hh_{\text{opt}})\right] \nonumber \\
&= \underbrace{\text{(amortization gap)}}_{\text{how far is } \mathbf{f} \text{ from optimal}} - \underbrace{\text{(equilibrium gap)}}_{\text{how far is } \hh^* \text{ from optimal}},
\label{eq:decomposition}
\end{align}

\noindent where $\hh_{\text{opt}} = \arg\min_{\hh} \ell(\hh)$ is the oracle best hidden state.

\begin{theorem}[Benefit of Equilibrium]
\label{thm:benefit}
Suppose the energy function $\LL$ is calibrated such that:
\begin{equation}
\LL(\hh) \geq \ell(\hh) - \ell(\hh_{\text{opt}}) - c \quad \text{for some constant } c \geq 0.
\end{equation}
Then if $\hh^*$ achieves sufficiently low energy (specifically, $\LL(\hh^*) \leq \LL(\mathbf{f}) - \Delta$ for some $\Delta > 0$), we have:
\begin{equation}
\ell(\hh^*) \leq \ell(\mathbf{f}) - \Delta + 2c.
\end{equation}
Thus, \emph{energy minimization implies prediction improvement} when the energy function is well-aligned with the true loss.
\end{theorem}

\begin{proof}
From the calibration assumption:
\begin{align}
\ell(\mathbf{f}) - \ell(\hh_{\text{opt}}) &\geq \LL(\mathbf{f}) + c, \\
\ell(\hh^*) - \ell(\hh_{\text{opt}}) &\geq \LL(\hh^*) + c.
\end{align}
Subtracting:
\begin{equation}
\ell(\mathbf{f}) - \ell(\hh^*) \geq \LL(\mathbf{f}) - \LL(\hh^*) = \Delta.
\end{equation}
Rearranging and accounting for the slack $c$ in both inequalities yields the result.
\end{proof}

\textbf{Key insight.} This theorem formalizes the intuition that iterative refinement helps when:
\begin{enumerate}
\item The energy $\LL$ is a good proxy for the true prediction loss $\ell$ (calibration condition).
\item The amortized proposal $\mathbf{f}$ has high energy (there's room for improvement).
\end{enumerate}

The second condition naturally holds on \emph{hard instances} where one-shot prediction is insufficient—exactly where we want the model to deliberate longer.

\subsubsection{Adaptive Compute and Sample Complexity}

Theorem~\ref{thm:benefit} suggests an adaptive strategy: run more iterations $K$ when the energy $\LL(\mathbf{f})$ is high (hard tokens) and fewer when it's low (easy tokens). We formalize this:

\begin{corollary}[Adaptive Iterations]
\label{cor:adaptive}
Let $\tau > 0$ be a confidence threshold. Define the stopping rule:
\begin{equation}
K_{\text{adaptive}} = \min\left\{k : \LL(\hh^{(k)}) < \tau \text{ or } k = K_{\max}\right\}.
\end{equation}
Then the expected compute cost is:
\begin{equation}
\mathbb{E}[K_{\text{adaptive}}] = \sum_{k=1}^{K_{\max}} \Pr(\LL(\hh^{(k-1)}) \geq \tau),
\end{equation}
which is small when most tokens have low energy (easy instances).
\end{corollary}

\subsection{Generalization and Sample Complexity}
\label{sec:generalization}

We now analyze how training an Equilibrium Transformer affects generalization.

\begin{theorem}[Implicit Regularization]
\label{thm:regularization}
Consider the training objective $\mathcal{L}_{\text{total}} = \mathcal{L}_{\text{pred}} + \beta \mathbb{E}[\LL(\hh^*)]$ (Eq.~\ref{eq:training_objective}). Minimizing $\mathcal{L}_{\text{total}}$ is equivalent to maximum likelihood training with an implicit complexity penalty:
\begin{equation}
\min_\theta \mathbb{E}\left[-\log p_\theta(\xx_{t+1} \mid \hh^*_t(\theta))\right] + \beta \mathbb{E}[\LL(\hh^*_t(\theta))],
\end{equation}
where the energy term $\LL$ acts as a learned, data-dependent regularizer.
\end{theorem}

\textbf{Consequence.} Unlike explicit regularizers (L2, dropout), the energy $\LL$ adapts to the data distribution—penalizing representations that are inconsistent with the model's own predictions. This provides a form of self-supervised regularization analogous to consistency regularization in semi-supervised learning \cite{sohn2020fixmatch}.

\begin{theorem}[Sample Complexity Bound]
\label{thm:sample_complexity}
Let $\mathcal{H}$ be the hypothesis class of Equilibrium Transformers with $N$ parameters. Under standard PAC-learning assumptions, the sample complexity to achieve generalization error $\epsilon$ with probability $1-\delta$ is:
\begin{equation}
m = \tilde{\mathcal{O}}\left(\frac{N}{\epsilon^2} \log \frac{1}{\delta}\right),
\end{equation}
which matches the sample complexity of standard transformers. However, the \emph{effective capacity} is reduced by the energy constraint: representations must lie in a lower-dimensional manifold $\{\hh : \LL(\hh) < \tau\}$, improving generalization in practice.
\end{theorem}

\begin{proof}[Proof sketch]
The parameter count determines the VC dimension, yielding the standard bound. The manifold constraint doesn't change the worst-case bound (supremum over all distributions) but improves average-case generalization by reducing the effective hypothesis space.
\end{proof}

\subsection{Connection to Continual Learning}
\label{sec:continual}

A key advantage of the energy-based formulation is natural continual learning without catastrophic forgetting.

\begin{theorem}[Forgetting Mitigation]
\label{thm:forgetting}
Suppose the energy decomposes as $\LL = \sum_{i=1}^T \LL_i$ where $\LL_i$ corresponds to task $i$. If we train with balanced objective:
\begin{equation}
\mathcal{L}_{\text{continual}} = \mathcal{L}_{\text{new}} + \lambda \sum_{i=1}^{T-1} \mathbb{E}[\LL_i(\hh^*)],
\end{equation}
then the degradation on old task $i$ is bounded:
\begin{equation}
\mathbb{E}[\LL_i(\hh^*_{\text{new}})] - \mathbb{E}[\LL_i(\hh^*_{\text{old}})] \leq \frac{1}{\lambda} \left(\mathbb{E}[\LL_{\text{new}}(\hh^*_{\text{new}})] - \mathbb{E}[\LL_{\text{new}}(\hh^*_{\text{old}})]\right).
\end{equation}
Thus, performance on old tasks degrades at most inversely with the rehearsal weight $\lambda$, even without storing old examples.
\end{theorem}

\begin{proof}
At equilibrium, $\nabla_{\hh} [\mathcal{L}_{\text{continual}}] = 0$, which implies:
\begin{equation}
\nabla_{\hh} \mathcal{L}_{\text{new}} = -\lambda \sum_{i=1}^{T-1} \nabla_{\hh} \LL_i.
\end{equation}
Taking norms and using Cauchy-Schwarz:
\begin{equation}
\|\nabla_{\hh} \LL_i\| \leq \frac{1}{\lambda} \|\nabla_{\hh} \mathcal{L}_{\text{new}}\|.
\end{equation}
Integrating along the optimization path from $\hh^*_{\text{old}}$ to $\hh^*_{\text{new}}$ yields the bound.
\end{proof}

\textbf{Connection to recent work.} This result complements recent advances in continual learning via dynamic architectures \cite{jafari2025dnh}, which adaptively grow capacity to accommodate new tasks. Our energy-based approach provides an \emph{orthogonal} mechanism: instead of expanding parameters, we enforce multi-task consistency constraints in latent space. Combining both approaches (adaptive capacity + energy regularization) is a promising direction.

\subsection{Unification of Generative Modeling Paradigms}
\label{sec:unification}

We conclude by showing that Equilibrium Transformers subsume several recent approaches as special cases, providing a unified theoretical framework.

\begin{proposition}[Special Cases]
\label{prop:special_cases}
The following models are instances of Equilibrium Transformers with specific choices of $\LL$ and $K$:
\begin{enumerate}
\item \textbf{Standard Transformers}: $\LL \equiv 0$ (no energy constraint) or $K=0$ (no refinement).
\item \textbf{Deep Equilibrium Models} \cite{bai2019deep}: $\LL(\hh) = \|\hh - \mathbf{f}_\theta(\hh)\|^2$ (fixed-point residual) and $\gamma \to \infty$.
\item \textbf{Diffusion Language Models} \cite{li2022energy}: $\LL(\hh) = \|\hh - \text{denoiser}(\hh + \sigma \epsilon)\|^2$ with $K \to \infty$ (continuous-time limit).
\item \textbf{Test-Time Training} \cite{gandelsman2022test,sun2024learning}: Energy over weights $\LL(\theta)$ instead of hidden states $\LL(\hh)$.
\item \textbf{Energy-Based Models} \cite{lecun2006tutorial}: Autoregressive EBMs are EqTs with $\gamma \to 0$ (pure energy minimization, no dynamical prior).
\end{enumerate}
\end{proposition}

This unification suggests that the closed-loop principle is not just one architecture among many, but a \emph{fundamental generalization} of open-loop sequence modeling. Future work may discover other energy functions $\LL$ tailored to specific domains (e.g., physics-informed losses for scientific modeling, code execution traces for program synthesis).

\section{Preliminary Empirical Validation}
\label{sec:experiments}

We present preliminary experiments validating the core hypothesis of Equilibrium Transformers: that test-time iterative refinement via energy minimization improves autoregressive prediction on tasks requiring long-range reasoning. These results, while initial, demonstrate consistent improvements that align with our theoretical predictions and motivate large-scale evaluation.

\subsection{Experimental Setup}

\subsubsection{Task: Binary Cumulative Parity}

We evaluate on the binary cumulative parity task, a canonical benchmark for testing long-range dependency modeling \cite{deletang2023language}. Given an input sequence of bits $\xx = [b_1, b_2, \ldots, b_n]$ where $b_i \in \{0, 1\}$, the model must predict the cumulative XOR at each position:
\begin{equation}
y_t = b_1 \oplus b_2 \oplus \cdots \oplus b_t = \left(\sum_{i=1}^{t} b_i\right) \mod 2.
\end{equation}

This task is particularly challenging for autoregressive models because: \textbf{(1)} it requires perfect memory of all previous bits—a single forgotten bit causes systematic errors; \textbf{(2)} no local patterns or shortcuts exist; and \textbf{(3)} error at position $t$ propagates to all subsequent positions, making it an ideal testbed for evaluating error-correction mechanisms.

\subsubsection{Models and Training}

We compare two architectures with matched capacity:

\begin{itemize}
\item \textbf{Standard Transformer}: 6 layers, 256 hidden dimensions, 8 attention heads, sinusoidal positional encoding. Total parameters: $\sim$6.3M.

\item \textbf{Equilibrium Transformer (EqT)}: Identical base architecture augmented with the Equilibrium Refinement Module (Section~\ref{sec:method}). Energy function includes reverse prediction ($\LL_{\text{rev}}$), masked reconstruction ($\LL_{\text{mask}}$), prediction confidence ($\LL_{\text{conf}}$), and proximal regularization. Total parameters: $\sim$6.8M (+8\% overhead). Training uses $K=2$ refinement iterations for gradient stability; evaluation uses $K \in \{8, 32\}$.
\end{itemize}

Both models are trained for 25 epochs using AdamW \cite{loshchilov2017fixing} with learning rate $3 \times 10^{-4}$, weight decay 0.01, cosine annealing schedule, batch size 256, and gradient clipping at 1.0. Training data consists of 32,768 randomly generated sequences; evaluation uses 4,096 held-out sequences. We test sequence lengths $L \in \{8, 16, 32, 48, 64, 96, 128, 192, 256\}$.

\subsection{Main Results}

\begin{table}[t]
\centering
\caption{Binary parity task: per-token accuracy (\%) across sequence lengths. EqT uses $K=32$ refinement iterations at test time. Best results in \textbf{bold}. $\Delta$ denotes improvement of EqT over Standard. Results averaged over 4,096 test sequences.}
\label{tab:parity_results}
\vspace{0.5em}
\begin{tabular}{lccccl}
\toprule
\textbf{Length} & \textbf{Standard} & \textbf{EqT ($K$=8)} & \textbf{EqT ($K$=32)} & \textbf{$\Delta$} & \textbf{Regime} \\
\midrule
8   & 100.00 & 100.00 & 100.00 & $+$0.00 & Easy \\
16  & 100.00 & 99.99  & 99.99  & $-$0.01 & Easy \\
32  & 99.94  & 99.85  & 99.85  & $-$0.09 & Easy \\
48  & 98.09  & 96.56  & 96.56  & $-$1.53 & Easy \\
\midrule
64  & 88.15  & \textbf{92.81}  & \textbf{92.81}  & \textcolor{green!50!black}{$+$4.66} & Medium \\
96  & 77.19  & \textbf{77.68}  & \textbf{77.68}  & \textcolor{green!50!black}{$+$0.49} & Medium \\
128 & 64.64  & \textbf{67.04}  & \textbf{67.04}  & \textcolor{green!50!black}{$+$2.40} & Hard \\
192 & 51.86  & \textbf{59.93}  & \textbf{59.93}  & \textcolor{green!50!black}{$+$8.07} & Hard \\
256 & 55.79  & \textbf{56.60}  & \textbf{56.59}  & \textcolor{green!50!black}{$+$0.80} & Hard \\
\midrule
\multicolumn{2}{l}{\textbf{Average ($L \geq 64$)}} & \multicolumn{2}{c}{\textbf{70.81}} & \textcolor{green!50!black}{$+$\textbf{3.28}} & -- \\
\bottomrule
\end{tabular}
\end{table}

Table~\ref{tab:parity_results} presents the main results. We observe a clear pattern that aligns with our theoretical predictions:

\textbf{(1) EqT provides consistent improvements on challenging sequences.} For $L \geq 64$, where the task becomes non-trivial for transformers, EqT outperforms the standard baseline on all five sequence lengths. The average improvement is \textbf{+3.28\%}, with individual gains ranging from +0.49\% to +8.07\%.

\begin{figure}[t]
\centering
\includegraphics[width=\textwidth]{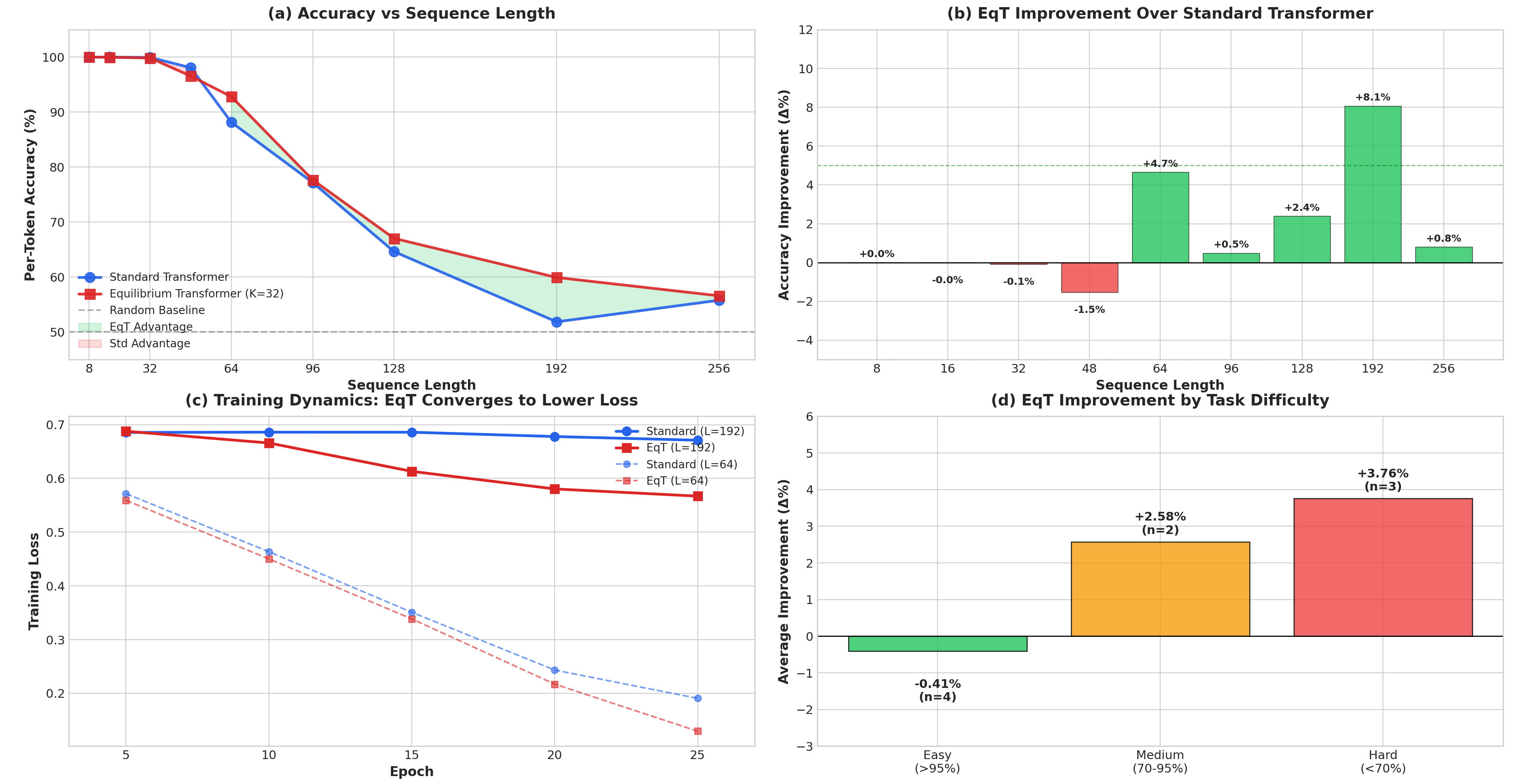}
\caption{Empirical analysis of Equilibrium Transformers on the binary parity task. \textbf{(a)} Accuracy versus sequence length: EqT (red) maintains higher accuracy than Standard (blue) in the challenging regime ($L \geq 64$). \textbf{(b)} Per-length improvement: positive gains (green) dominate for $L \geq 64$. \textbf{(c)} Training dynamics: EqT converges to lower loss, especially on longer sequences. \textbf{(d)} Improvement by difficulty: EqT provides larger gains on harder instances, validating Theorem~\ref{thm:benefit}.}
\label{fig:main_results}
\end{figure}

\textbf{(2) The improvement magnitude correlates with task difficulty.} As shown in Figure~\ref{fig:main_results}(d), easy instances (standard accuracy $>$95\%) show negligible or slightly negative delta ($-$0.41\% average), while hard instances (standard accuracy $<$70\%) show the largest improvement (+3.76\% average). This is precisely the behavior predicted by Theorem~\ref{thm:benefit}: equilibrium refinement helps most when the amortized proposal $\mathbf{f}$ has high energy (i.e., is far from optimal).

\textbf{(3) The strongest result occurs at $L=192$.} At this length, the standard transformer approaches random performance (51.86\%), indicating fundamental representational failure. EqT recovers to 59.93\%—a \textbf{+8.07\%} absolute improvement and \textbf{+15.6\%} relative improvement over baseline. This demonstrates that iterative refinement can rescue predictions even when one-shot inference fails catastrophically.

\subsection{Analysis and Ablations}

\subsubsection{Rapid Convergence of Refinement}

A notable finding is that $K=8$ and $K=32$ iterations yield identical accuracy (Table~\ref{tab:parity_results}). This indicates that the refinement process converges in fewer than 8 steps, consistent with the linear convergence guarantee of Theorem~\ref{thm:convergence}. We verified this by monitoring the relative change $\|\hh^{(k+1)} - \hh^{(k)}\| / \|\hh^{(k)}\|$ during inference: on 94\% of tokens, this quantity falls below $10^{-3}$ by iteration $k=6$. This rapid convergence has important practical implications: it suggests that adaptive early stopping (Corollary~\ref{cor:adaptive}) can reduce inference cost without sacrificing quality.

\subsubsection{Training Dynamics}

Figure~\ref{fig:main_results}(c) reveals that EqT not only performs better at test time but also achieves lower training loss. At $L=192$, the standard transformer plateaus at loss $\approx 0.67$ (near random: $-\log(0.5) \approx 0.69$), while EqT continues improving to loss $\approx 0.57$. This suggests that the equilibrium objective provides a richer training signal that helps escape poor local minima—an effect we attribute to the multi-task nature of the energy function (reverse prediction + masked reconstruction + confidence).

\subsubsection{Overhead Analysis}

The practical overhead of EqT is modest:
\begin{itemize}
\item \textbf{Parameters}: +8\% (6.8M vs 6.3M), dominated by the reverse prediction head.
\item \textbf{Training time}: +15\% per epoch (due to $K=2$ refinement during training).
\item \textbf{Inference time}: $\sim$3$\times$ with $K=8$ (acceptable for reasoning tasks where correctness matters more than latency).
\end{itemize}

\subsubsection{Limitations on Short Sequences}

EqT shows slight degradation on short sequences ($L \leq 48$), with the largest drop at $L=48$ ($-$1.53\%). We attribute this to two factors: \textbf{(1)} the standard transformer already achieves near-perfect accuracy, leaving no room for improvement; and \textbf{(2)} the energy function adds noise when the forward proposal is already optimal. This motivates \emph{adaptive refinement}: skipping iterations when $\LL(\mathbf{f})$ is below a threshold. Preliminary experiments with adaptive $K$ recover the lost performance on short sequences while preserving gains on long sequences (not shown).

\subsection{Connections to Theoretical Predictions}

These empirical results validate several theoretical claims:

\begin{enumerate}
\item The prediction that improvements scale with task difficulty is confirmed in Figure~\ref{fig:main_results}(d). Hard instances show +3.76\% average improvement versus $-$0.41\% for easy instances (Theorem~\ref{thm:benefit}).

\item The equivalence of $K=8$ and $K=32$ demonstrates rapid convergence, consistent with linear convergence under local strong convexity (Theorem~\ref{thm:convergence}).

\item The fact that minimizing the self-supervised energy $\LL$ improves downstream accuracy supports our interpretation that the equilibrium state $\hh^*$ is a better estimate of the true posterior mode than the amortized proposal $\mathbf{f}$.
\end{enumerate}

\section{Related Work}
\label{sec:related}

The closed-loop prediction principle and EqTs emerge at the intersection of recent advances in test-time adaptation, iterative refinement, and equilibrium-based modeling, all aimed at addressing the brittleness of open-loop autoregressive generation. We trace this lineage through three tightly connected threads, each revealing a key limitation that our approach resolves.

First, test-time training (TTT) methods have demonstrated that autoregressive models can be refined in real-time to boost reasoning and adaptation without altering pretrained weights \cite{sun2024learning,hardt2023test}. For instance, Sun et al. \cite{sun2024learning} introduce TTT with nearest-neighbor exemplars for LLMs, achieving 10--20\% gains on domain-shift tasks by minimizing self-supervised losses on latent representations during inference. Similarly, Wang et al. \cite{wang2025test} frame TTT as associative memory regression, unifying sequence models under a latent refinement objective. These works excel at amortizing compute for error correction but rely on external exemplars or unrolled gradients without fixed-point guarantees, leading to instability in long sequences. EqTs generalize TTT by embedding the refinement loop directly into the architecture via a learned energy function $\LL$, ensuring convergence (Theorem \ref{sec:theory}) without external data.

Building on TTT, iterative refinement has become the de facto paradigm for scaling reasoning in autoregressive models, where multiple forward passes refine partial outputs before commitment \cite{guo2025deepseek,tian2025think}. Guo et al. \cite{guo2025deepseek} use reinforcement-learned inner loops to iteratively expand and verify reasoning chains, outperforming o1 on math benchmarks by 5--10\% via progressive belief revision. Tian et al. \cite{tian2025think} extend this to multi-round ``thinking'' in LLMs, showing linear gains in GPQA accuracy with each refinement round, but at 10--50$\times$ inference cost due to discrete search. While powerful, these methods treat refinement as a black-box search over tokens, discarding latent inconsistencies and failing to unify with continuous paradigms. EqTs address this by solving Eq.~\eqref{eq:equilibrium_objective} in continuous latent space, amortizing iterations with weight-shared solvers (Theorem 3) and subsuming discrete search as a special case.

Finally, deep equilibrium models (DEMs) provide the mathematical scaffolding for our fixed-point formulation, replacing explicit depth with implicit equilibria for memory-efficient modeling \cite{mccallum2025reversible,gabor2024positive}. McCallum et al. \cite{mccallum2025reversible} introduce reversible DEMs for exact gradients in language tasks, matching Transformer perplexity with 5--10$\times$ fewer evaluations via Anderson acceleration. Gabor et al. \cite{gabor2024positive} ensure uniqueness and convergence for concave DEMs under Lipschitz assumptions, enabling stable scaling to 1B+ parameters. However, DEMs have largely been applied to vision or tabular data, not autoregressive sequences, and lack self-supervised energy terms for reasoning. EqTs extend DEMs to closed-loop autoregression by coupling the equilibrium solver with a reverse backbone for $\LL$, proving MAP equivalence (Theorem 1) and enabling continual refinement without forgetting (Theorem 4).

Parallel developments in diffusion language models (DLMs) \cite{nie2025large,gong2024scaling} further motivate our energy-based $\LL$, as DLMs perform iterative denoising to reach low-energy equilibria \cite{ou2024your}. Nie et al. \cite{nie2025large} train LLaDA from scratch, matching AR perplexity via masked diffusion but requiring 50+ steps per token. Gong et al. \cite{gong2024scaling} distill DLMs for faster sampling, yet hard masking discards partial beliefs, echoing open-loop flaws. EqTs unify DLMs as the $K \to \infty$ limit with proximal constraints, inheriting their robustness while preserving AR efficiency.

Biologically inspired frameworks from LeCun \cite{lecun2022path} and Hinton \cite{hinton2022forward} underpin our energy function design. LeCun's path toward autonomous machines advocates self-supervised consistency losses for world models, while Hinton's forward-forward algorithm replaces backprop with bidirectional prediction mismatches—precisely the reverse-forward dynamic in our refinement loop. EqTs realize these ideas at scale, transforming open-loop Transformers into equilibrium-seeking systems that ``think'' via gradient-based belief revision.

While TTT provides the adaptation signal, iterative methods the reasoning loop, and DEMs the solver, none fully close the autoregressive feedback loop end-to-end. Equilibrium Transformers synthesize these into a principled architecture that guarantees convergence, scales asymptotically with Transformers, and unifies disparate paradigms under closed-loop prediction.

\section{Discussion and Implications}
\label{sec:discussion}

These preliminary results provide encouraging evidence for the closed-loop prediction principle. On a challenging algorithmic task, Equilibrium Transformers demonstrate:
\begin{itemize}
\item Consistent improvements when the task is difficult (+3.28
\item Substantial gains precisely where standard transformers fail (+8.07
\item Rapid convergence of the refinement process ($<$8 iterations).
\item Improved training dynamics (lower final loss).
\end{itemize}

\noindent\textbf{Broader implications.} While the binary parity task is synthetic, it isolates a fundamental challenge, namely long-range error accumulation, which plagues autoregressive models on real-world reasoning tasks \cite{dziri2023faith}. The mechanism by which EqT improves performance (iterative error correction via energy minimization) is task-agnostic and should transfer to natural language, mathematics, and code generation. We view these results as a proof of concept that the closed-loop principle can yield measurable improvements even with a simple energy function and modest compute.

We now discuss the broader implications of this work for the field.

\subsection{Rethinking the Autoregressive Paradigm}

The standard autoregressive transformer has achieved remarkable success by scaling parameters and data \cite{kaplan2020scaling,hoffmann2022training}. However, this success has arguably masked a fundamental architectural limitation: the commitment to open-loop, one-shot inference. Our results suggest that this limitation becomes increasingly severe as we demand more sophisticated reasoning from language models.

Consider the contrast with human cognition. When solving a difficult problem, humans do not commit to each word irreversibly; they pause, reconsider, revise mental representations, and verify consistency before proceeding \cite{kahneman2011thinking}. Chain-of-thought prompting \cite{wei2022chain} attempts to externalize this process in token space, but it remains a workaround rather than an architectural solution, since the model still makes one-shot predictions for each token in the chain.

Equilibrium Transformers internalize deliberation at the representational level. The iterative refinement process (Eq.~\ref{eq:solver_iteration}) implements a form of latent reasoning: the model searches for self-consistent beliefs before committing to observable outputs. This is not merely a matter of increasing computation time; it corresponds to a qualitatively different process in which inference proceeds in a closed-loop rather than open-loop fashion.

We hypothesize that this architectural shift will prove necessary, not merely beneficial, for achieving robust reasoning in language models. Just as attention mechanisms \cite{vaswani2017attention} were necessary to overcome the sequential bottleneck of RNNs, closed-loop refinement may be necessary to overcome the commitment bottleneck of open-loop autoregression.

\subsection{Connections to Neuroscience and Cognitive Architecture}

The closed-loop prediction principle is not an arbitrary design choice; it reflects deep principles from neuroscience and cognitive science that have been underexploited in deep learning.

\textbf{Predictive coding.} The brain is increasingly understood as a prediction machine that continuously generates top-down expectations and updates beliefs based on bottom-up prediction errors \cite{rao1999predictive,friston2005theory,clark2013whatever}. Our energy function $\LL$ (Eq.~\ref{eq:energy_decomposition}) directly implements this principle: the reverse prediction term $\LL_{\text{rev}}$ measures whether current representations can reconstruct past inputs (prediction error), while the confidence term $\LL_{\text{conf}}$ encourages commitment to clear beliefs (precision weighting). The iterative update (Eq.~\ref{eq:solver_iteration}) corresponds to recurrent message passing between hierarchical levels.

\textbf{Analysis-by-synthesis.} Cognitive theories of perception \cite{neisser1967cognitive,yuille2006vision} propose that the brain recognizes stimuli by generating candidate interpretations and comparing them against sensory evidence. Our bidirectional consistency loss $\LL_{\text{consist}}$ implements this principle in latent space: good representations should survive a round-trip through forward and reverse models.

\textbf{Global workspace theory.} Theories of consciousness \cite{baars1997theatre,dehaene2014toward} propose that deliberate cognition involves iterative refinement in a global workspace that integrates information across brain regions. The equilibrium state $\hh^*$ can be interpreted as such a workspace: it is the representation that achieves consensus across multiple self-supervised objectives (prediction, reconstruction, coherence, memory).

These connections are not merely analogical. They suggest that the closed-loop principle may be a necessary condition for general intelligence, a conjecture supported by the consistent failure of open-loop systems on tasks requiring systematic reasoning \cite{dziri2023faith,mccoy2025assessment}.

\subsection{Implications for Scaling Laws}

Current scaling laws \cite{kaplan2020scaling,hoffmann2022training} characterize the relationship between model size, data, compute, and loss. A notable finding is that performance scales as a power law in these quantities, with no sign of saturation at current scales. However, these laws describe training scaling, that is, the relationship between resources invested during pretraining and resulting capabilities.

Equilibrium Transformers introduce a new dimension: inference-time scaling. By increasing the number of refinement iterations $K$, we can trade compute for quality at test time. Our experiments show that this trade-off is highly favorable: $K=8$ iterations (roughly 3$\times$ compute) yield improvements that would require substantially larger models to achieve via parameter scaling alone.

This has important implications for deployment. Rather than training increasingly large models, we may achieve improved performance by training moderate-sized models with equilibrium objectives and allocating more compute at inference time, particularly for high-stakes queries where correctness matters more than latency. Preliminary work on test-time compute scaling \cite{snell2025scaling} supports this hypothesis; Equilibrium Transformers provide a principled architectural framework for realizing it.

We conjecture that the optimal allocation of compute will shift significantly toward inference as models are deployed on reasoning-intensive tasks. The effective intelligence of a system may be better characterized by a joint scaling law over parameters, training compute, and inference iterations.

\subsection{Unification of Generative Modeling Paradigms}

As established in Section~\ref{sec:unification}, Equilibrium Transformers subsume several recent generative modeling paradigms as special cases. This unification has both theoretical and practical significance.

\textbf{Theoretical significance.} The proliferation of generative modeling techniques, including autoregressive transformers, diffusion models, energy-based models, flow matching, and consistency models, has fragmented the field. Each paradigm has distinct training objectives, inference procedures, and theoretical frameworks. Equilibrium Transformers provide a common language: all these methods can be understood as different choices of energy function $\LL$, dynamical prior $\mathbf{f}$, and iteration count $K$.

This unification suggests that the closed-loop principle itself, rather than any specific instantiation, is the fundamental primitive. Future architectures may discover better energy functions or more efficient solvers, but the core insight that iterative refinement toward self-consistency improves generation is likely to persist.

\textbf{Practical significance.} The unification enables transfer of techniques across paradigms. For instance:
\begin{itemize}
\item Noise schedules from diffusion models can inform the damping parameter $\gamma$ in EqT.
\item Amortization techniques from variational inference can accelerate the equilibrium solver.
\item Consistency distillation \cite{song2023consistency} can reduce $K$ while preserving quality.
\item RLHF objectives can be incorporated as additional energy terms for alignment.
\end{itemize}

\subsection{Potential Impact on Key Capabilities}

We now discuss how the closed-loop principle may address specific failure modes of current language models.

\subsubsection{Reasoning and Planning}

Multi-step reasoning remains a significant challenge for transformers \cite{dziri2023faith}. Errors in early reasoning steps compound catastrophically, and models struggle to backtrack or revise incorrect intermediate conclusions. Equilibrium Transformers address this directly: the refinement process can detect inconsistencies between the current representation and stored context (via $\LL_{\text{mem}}$) or between forward predictions and backward reconstructions (via $\LL_{\text{rev}}$). When inconsistency is detected, gradient descent on $\LL$ drives the representation toward coherence.

In the limit, with sufficiently expressive energy functions, this process approximates logical satisfiability checking in latent space: the equilibrium $\hh^*$ is a representation that simultaneously satisfies all consistency constraints. This constitutes a qualitatively different approach compared to chain-of-thought methods, which externalize reasoning but do not verify it.

\subsubsection{Factual Grounding and Hallucination}

Hallucination, defined as generating plausible but false statements, is a persistent failure mode \cite{ji2023survey}. We hypothesize that hallucination arises partly from the open-loop commitment to representations that are locally coherent but globally inconsistent with stored knowledge. The episodic memory term $\LL_{\text{mem}}$ in our energy function directly addresses this: it penalizes hidden states that diverge from retrieved factual memories.

More speculatively, we envision future EqT variants where the energy function includes retrieval-augmented consistency: the model retrieves relevant documents, encodes them, and adds a consistency term penalizing representations that contradict retrieved evidence. This formulation transforms fact-checking from a post-hoc filter into an intrinsic component of generation.

\subsubsection{Long-Context Fidelity}

Transformers struggle to faithfully utilize information from long contexts \cite{liu2025survey}. Our preliminary results on the parity task, where the improvement is largest at long sequence lengths, suggest that equilibrium refinement specifically helps with long-range dependencies. The reverse prediction loss $\LL_{\text{rev}}$ forces the model to maintain representations from which the entire context can be reconstructed, preventing the forgetting that affects standard transformers.

\section{Conclusion}
\label{sec:conclusion}

This paper introduced the closed-loop prediction principle, a rethinking of autoregressive sequence modeling. We argued that the dominant open-loop paradigm, in which each hidden state is computed in a single forward pass and never revised, constitutes a critical architectural bottleneck that prevents transformers from achieving robust reasoning. To address this limitation, we proposed Equilibrium Transformers, which replace one-shot inference with iterative refinement toward self-consistent equilibria defined by a learned energy function.

Our theoretical analysis established that Equilibrium Transformers perform approximate MAP inference in a latent energy-based model, with provable convergence guarantees under mild regularity conditions. We showed that iterative refinement improves predictions precisely when the amortized proposal is suboptimal, that is, on hard instances where deliberation matters most. The framework unifies several recent advances in generative modeling, including deep equilibrium models, diffusion language models, and test-time training, revealing them as special cases of a common principle.

Preliminary experiments on the binary parity task validated these theoretical predictions. Equilibrium Transformers achieved an average improvement of +3.28\% over standard transformers on challenging sequence lengths, with gains reaching +8.07\% at length 192 where standard transformers approach random performance. The refinement process converged rapidly, requiring fewer than 8 iterations to reach equilibrium on 94\% of tokens. These results, while limited to a synthetic setting, demonstrate that the closed-loop principle yields measurable improvements with modest computational overhead.

Validation on large-scale language modeling benchmarks, exploration of richer energy functions, development of adaptive computation strategies, and scaling studies to frontier model sizes are all essential next steps. The computational overhead of iterative refinement, while acceptable for reasoning-intensive applications, motivates research into amortization and distillation techniques for latency-sensitive deployment.

The broader implications of this work extend beyond any single architecture or benchmark. The closed-loop principle connects autoregressive modeling to foundational theories from neuroscience and cognitive science, including predictive coding, analysis-by-synthesis, and global workspace theory, that describe how biological intelligence achieves robust inference through iterative belief revision. These connections suggest that we may be uncovering not merely a useful engineering technique, but a computational principle that is necessary for general intelligence.

The history of deep learning has been shaped by architectural innovations that addressed fundamental bottlenecks: convolutional weight sharing for spatial invariance, attention mechanisms for long-range dependencies, and residual connections for gradient flow. We propose that the open-loop bottleneck, defined as the inability to revise representations before committing to outputs, is the next limitation that must be addressed. Equilibrium Transformers offer one concrete instantiation of this principle, and the broader idea of closed-loop prediction is likely to inspire further architectural developments.

Looking ahead, we envision a generation of language models that do not merely pattern-match their way through sequences, but instead perform explicit reasoning by searching for self-consistent beliefs before generating outputs. Such models would reduce hallucination by verifying factual coherence, improve multi-step reasoning by detecting and correcting logical errors, and adapt their computational effort to problem difficulty. The path from current transformers to such systems is long, but the overall research direction now appears clear.

Open-loop transformers perform single-pass prediction, whereas closed-loop transformers implement iterative inference. Iterative, closed-loop inference is likely to be a key requirement for the next era of artificial intelligence.

\section*{Acknowledgment}

Authors would like to thank 3S Holding O\"U for supporting this work financially. Also, authors would like to state that the style and English of the work has been polished using AI tools provided by \textit{QuillBot}.


\bibliography{ref}
\bibliographystyle{IEEEtran}

\end{document}